\documentclass[a4paper,twoside]{article}
\usepackage{epstopdf}
\usepackage{epsfig}
\usepackage{subcaption}
\usepackage{calc}
\usepackage{amssymb}
\usepackage{amstext}
\usepackage{amsmath}
\usepackage{amsthm}
\usepackage{multicol}
\usepackage{pslatex}
\usepackage{apalike}
\usepackage{graphicx}
\usepackage{url}
\usepackage{SCITEPRESS}     % Please add other packages that you may need BEFORE the SCITEPRESS.sty package.

\begin{document}

\title{Real-time Detection of 2D Tool Landmarks with Synthetic Training Data}

\author{\authorname{Bram Vanherle\sup{1}, Jeroen Put\sup{1}, Nick Michiels\sup{1} and Frank Van Reeth \sup{1}}
\affiliation{\sup{1}Hasselt University - tUL - Flanders Make, Expertise Centre for Digital Media\\
  Agoralaan, 3590 Diepenbeek, Belgium}
\email{firstname.lastname@uhasselt.be}
}


\abstract{In this paper a deep learning architecture is presented that can, in real time, detect the 2D locations of certain landmarks of physical tools, such as a hammer or screwdriver. To avoid the labor of manual labeling, the network is trained on synthetically generated data. Training computer vision models on computer generated images, while still achieving good accuracy on real images, is a challenge due to the difference in domain. The proposed method uses an advanced rendering method in combination with transfer learning and an intermediate supervision architecture to address this problem. It is shown that the model presented in this paper, named Intermediate Heatmap Model (IHM), generalizes to real images when trained on synthetic data. To avoid the need for an exact textured 3D model of the tool in question, it is shown that the model will generalize to an unseen tool when trained on a set of different 3D models of the same type of tool. IHM is compared to two existing approaches to keypoint detection and it is shown that it outperforms those at detecting tool landmarks, trained on synthetic data.}

\onecolumn \maketitle \normalsize \setcounter{footnote}{0} \vfill

\section{\uppercase{Introduction}}
\label{sec:introduction}
\begin{figure}
    \centering
    \includegraphics[width=\linewidth]{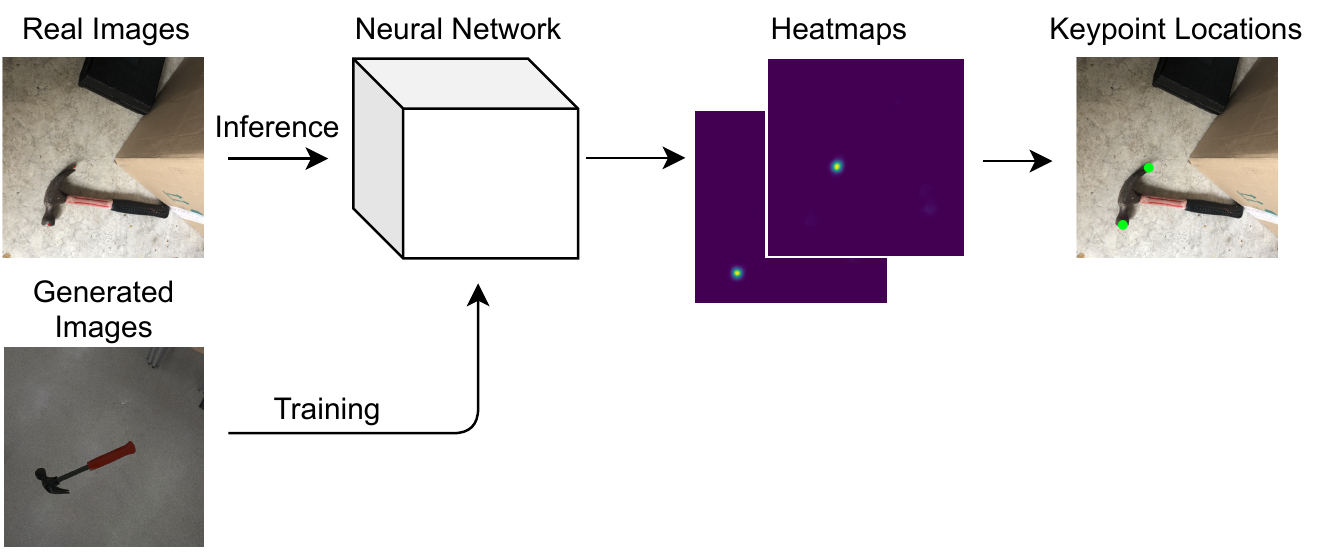}
    \caption{An overview of the proposed method for keypoint localization.}
    \label{fig:overview}
\end{figure}

The modern manufacturing industry requires more intricate technical knowledge from the operators assembling products, as those products become more complex. This in turn, creates a need for more advanced vision systems to both support and train the humans assembling these products. The goal of this paper is to detect the 2D location of certain landmarks in an RGB image of a tool being used in such assembly processes. This could, for example, be the head of a  screwdriver or the two ends of a combination wrench. There are a lot of variables--such as pose, occlusions, lighting and physical variations--that impact the appearance of a tool in an image. This makes it challenging to robustly localize that tool's landmarks.\\
The developed method has a strong focus on execution speed so that it can be used in real-time video analysis systems. An example application is the automatic monitoring of assembly progress, where our method can be used to detect if tool-based actions are completed. This could be used to offer the operator adaptive feedback. In this paper the method is applied to assembly tools, but this work could easily generalize to different types of objects.\\
Deep Convolutional Neural Networks have proven to be an exceptional tool at interpreting visual information due to their ability to automatically detect features at multiple levels. An important drawback of using deep learning is that large amounts of labeled data are required. To ensure this system is easily adopted to new types of tools, the network is trained solely on computer generated images. This avoids a human having to make pictures of tools and manually label these to apply this method to a different tool. Whereas the usage of synthetic data allows for easy collection of data, it often leads to a decrease in accuracy. This is due to the model being trained on a different domain than it is being used on during inference. The negative impact of the domain gap is mitigated by using a specialized network architecture, and data generation techniques that are optimized for deep learning input. A full overview of method presented in this paper is shown in Figure~\ref{fig:overview}\\
This paper proposes a Fully Convolutional Neural Network that generates heatmaps indicating the probability that a certain landmark is at the specified location. The novel Intermediate Heatmap Model (IHM) combines a feature detection network with upsampling layers. Intermediate supervision is implemented by generating output at multiple levels to boost the learning of higher level features. To prove the usability of this method, it is compared to two existing keypoint detection methods on a custom tool dataset. The method is also validated on the Pascal3D~\cite{pascal3d} dataset. Our main contributions are:
\begin{itemize}
    \item A method for quickly generating images that are suitable for deep learning, along with appropriate labels.
    \item A Deep Learning architecture for keypoint detection that achieves good performance when trained on synthetic images.
\end{itemize}

\section{Related Work}
In classic computer vision, keypoint detection is usually done using feature descriptors such as SIFT~\cite{SIFT} or SURF~\cite{SURF}. These methods focus on local features and fail to consider the global context of the image, hence why they do not generalize well to unseen data and fail under variations of pose, lighting and texture.\\
In more recent works there has been a shift towards Deep Learning to avoid these issues. Early works, such as DeepPose~\cite{DeepPose}, would use a CNN to detect image features, followed by fully connected layers that directly regress the $(x, y)$ coordinates of the keypoints. These methods had low accuracy, which lead to the creation of networks that regress heatmaps for each keypoint. Keypoint locations are computed from the maxima of these heatmaps. This method has been used for 6DoF Pose Estimation~\cite{semantic_keypoints}, Object Detection~\cite{corner_net} and mainly for Human Pose Estimation~\cite{pose_machines,stacked_hourglass,simple_baselines}. These methods are easier to train, but have difficulties detecting keypoints when objects are occluded or truncated. For this reason Pixelwise-Voting network~\cite{pvnet} was introduced. This type of network learns, for each keypoint, to generate unit vectors at each pixel pointing to that keypoint. Those unit vectors are then used to vote for keypoint locations using RANSAC~\cite{ransac}. This forces the network to focus on local features and spatial relations between object parts, and allows it to infer the position of invisible parts. \\
The method proposed in this paper falls under the category of heatmap generating models. Although the pixel-wise voting method can handle occlusions better, the RANSAC voting adds extra computational complexity at runtime which is to be avoided in a real-time system. Existing heatmap models often have fast inference times, but rely on real images with annotations as training data. Our method removes this need by offering a data generation method and a deep learning model that performs well, when trained on synthetic images.

\section{Training Data} \label{sec:training_data}
\subsection{Image Generation}
\begin{figure}
\centering
\includegraphics[width = 0.3\linewidth]{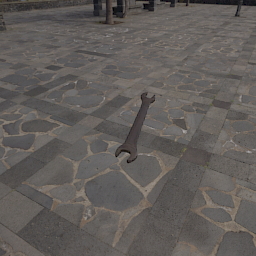}
\includegraphics[width = 0.3\linewidth]{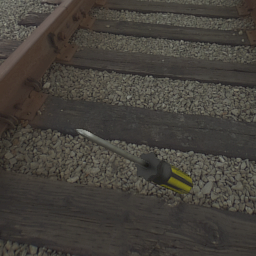}
\includegraphics[width = 0.3\linewidth]{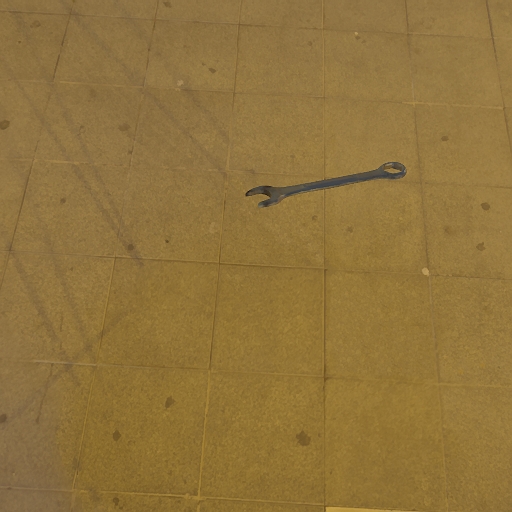}
\includegraphics[width = 0.3\linewidth]{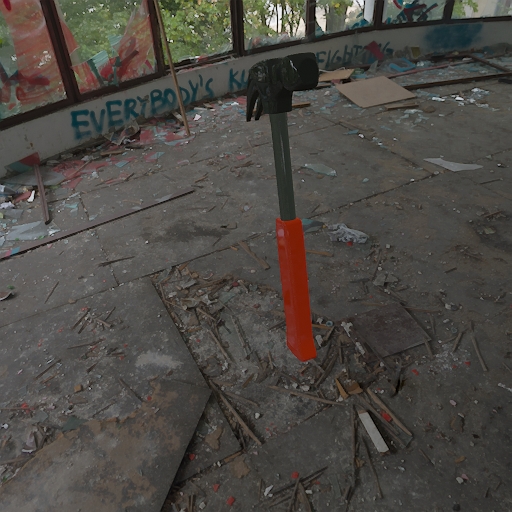}
\includegraphics[width = 0.3\linewidth]{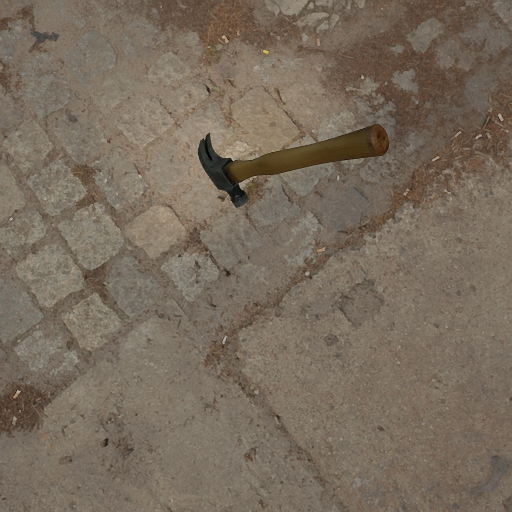}
\includegraphics[width = 0.3\linewidth]{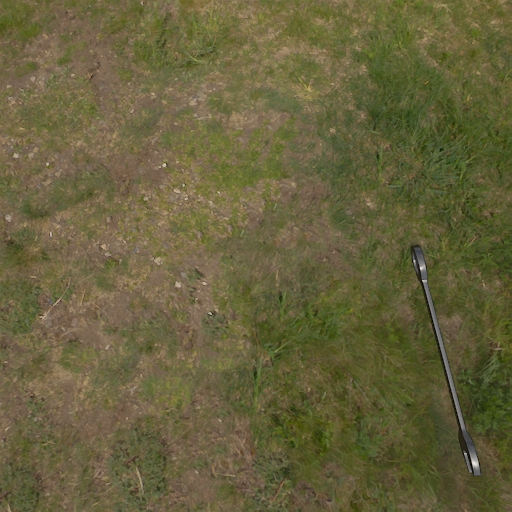}
\caption{Synthetically generated images of tools.\label{fig:syntheticData}}
\end{figure}

A key challenge of using a data-driven approach is data collection. Training a model for a real-time 2D landmark detection requires the collection of sufficient training data with a broad range of variations for each physical tool at hand. Such training datasets are not easily available and building new ones requires access to large amounts of representative existing image data. Additionally, manual labor for identifying the ground truth 2D landmarks is required, which is cumbersome and introduces human error to the dataset. Furthermore, this approach does not generalize well when new types of physical tools are introduced. \\
To avoid the time consuming process of taking and manually labeling thousands of photos of tools, we propose a synthetic data generation tool to easily generate training sets of newly introduced tools. Synthetic data generation is often done by using a Cut and Paste strategy~\cite{cut_and_paste,simple_copy_paste}. This entails cutting the foreground objects from a limited set of images and pasting these at randomized locations on random backgrounds. This method lends itself well to Object Detection and Segmentation problems, as those labels can be directly inferred from the image composition. Although this method can quickly generate large datasets, the images produced are often not realistic and lack variation in appearance, as the foreground objects are typically sourced from a very limited set. This could have a negative impact on the accuracy of the keypoint localization. We therefore propose a tool that can render a set of images by constructing fully random 3D scenes with the desired objects in them.\\
This tool will allow for more intricate appearance and 3D variations than the classic data augmentation techniques such as noise and translation. Examples of more representative variations are 3D physical tool rotations, translations, camera pose, occlusions, shadows, lighting conditions, etc. Furthermore, we aim to generate images as photorealistically as possible, as this has been shown to improve performance on other deep learning tasks~\cite{photorealism}. To achieve this, we have exploited the real-time raytracing capabilities of recent NVidia RTX hardware.\\
To be able to create a scene with the desired object in it, a 3D model of the object is needed. These can be acquired by either performing a 3D scan or by manually modeling the object in CAD software. Both these options are time consuming and can introduce errors. We argue that this can be avoided by using a set of 3D models of variations on the object, as opposed to one exact model of the object. These variations can differ in shape or texture, as long as they portray the same type of object. By training a neural network on a set of tools that all sufficiently vary in appearance, the network will generalize to unseen tools of the same kind. This allows us to simply source our 3D models from the internet, for example.\\
The main input of this tool is a set of textured 3D meshes of the physical tools. The tool will automatically generate a wide range of variations based on these models. The following parameters are randomized:
\begin{itemize}
    \item Object position
    \item Object rotation
    \item Environment
    \item Lighting
\end{itemize}
Each random sample will be rendered with a path tracer to create a photo-realistic image. Figure~\ref{fig:syntheticData} shows different examples of randomly generated synthetic image data. An advantage of using such a synthetic data generator is that we can export any kind of additional labeled metadata as well. Here, the ground truth 2D positions of the landmarks are exported together with the rendered images. This results in an automatic approach for generating large amounts of data needed for training deep learning networks  such as the one presented in Section~\ref{sec:architecture}. The use of path tracing ensures a sufficient amount of visual realism. \\
To further increase the size of the dataset and to add occlusions, a copy of each image is added to the dataset with a random object pasted in front of it, or with the background switched out for a random one. Foregrounds and backgrounds are added to the image using a random blending method to avoid the network focusing on pixel artifacts, as described by \cite{cut_and_paste}. Blending was either done by alpha blending, Poisson blending~\cite{poisson_blend} or Laplacian pyramid blending~\cite{laplacian_blending}.

\subsection{Heatmaps}
\begin{figure}
    \centering
    \includegraphics[width = \linewidth]{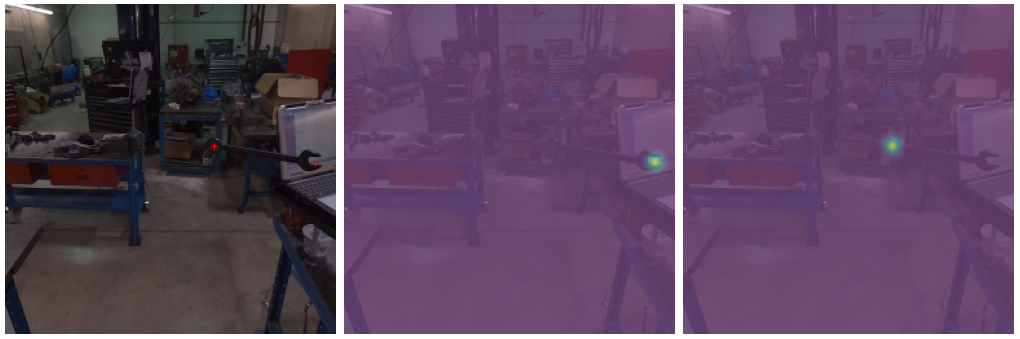}
    \caption{Synthetically generated image of a combination wrench and the two heatmaps that the neural network will try to predict.}
    \label{fig:wrench_hm}
\end{figure}

In the proposed method, a Convolutional Neural Network is used to interpret the image. This creates a feature map. The locations of the keypoints need to then be extracted from this feature map. This could be done by flattening the features and using one or more fully connected layers to transform that vector into the coordinate values of the keypoints. The flattening operation obfuscates the spatial information, and the fully connected layers, unlike convolutional layers, are not spatially invariant. This adds learning complexity that leads to lower accuracy. A better approach is to use a fully convolutional neural network that generates a set of heatmaps, as opposed to numerical coordinates. A heatmap is generated for each keypoint and contains, for each pixel, the probability that the keypoint is at that location. Such a set of heatmaps can be derived from the feature map by using convolutions, which uses the spatial information more efficiently.\\
To construct the training data for an image a heatmap is generated for each keypoint. A heatmap is created by placing a two-dimensional Gaussian kernel, with $\sigma = \frac{1}{64} \cdot \textrm{width}$, centered at the 2D image location of the keypoint. A smaller value for $\sigma$ would make it difficult for the network to learn, whereas a higher value would decrease localization accuracy. Using this method, the value of the heatmap is $1.0$ at the keypoint location, with values tapering off further away from the keypoint. An alternative to this is to only make the correct pixel $1.0$, but this is difficult to learn due to the overwhelming amount of zero values. Assigning $1.0$ to a larger area around the ground truth pixel would solve this, but this would make it impossible to retrieve the exact pixel location. A Gaussian kernel is therefore ideal in this scenario. Figure~\ref{fig:wrench_hm} shows the two Gaussian heatmaps generated for the two keypoints of a combination wrench. When keypoints are very similar, as is the case for the two ends of an open ended wrench, the heatmaps are added together into one heatmap. This removes ambiguity, making learning easier.\\
During inference, the maximum of each heatmap is computed to find the pixel location of that keypoint. When the heatmaps are compressed into one heatmap, for the case of similar keypoints, local maxima are then used as keypoint locations instead of the global maximum. The value of the maximum pixel can be interpreted as the confidence of the prediction, which can be used to threshold less certain predictions.

\section{Method}
\subsection{Network Architecture} \label{sec:architecture}  
\begin{figure}
    \centering
    \includegraphics[width=\linewidth]{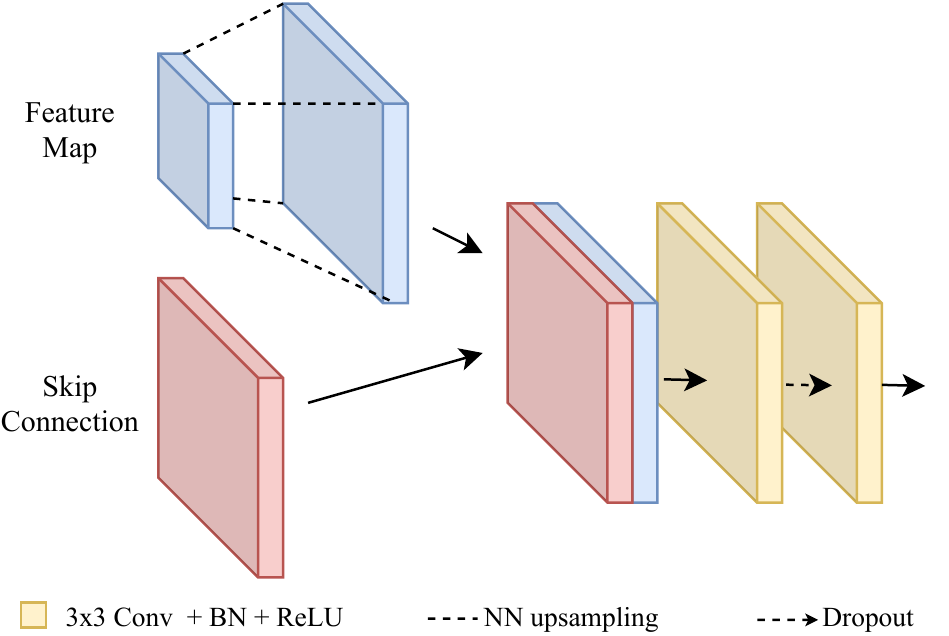}
    \caption{Illustration of the upsampling block.}
    \label{fig:upsampling}
\end{figure}

\begin{figure}
    \centering
    \includegraphics[width=\linewidth]{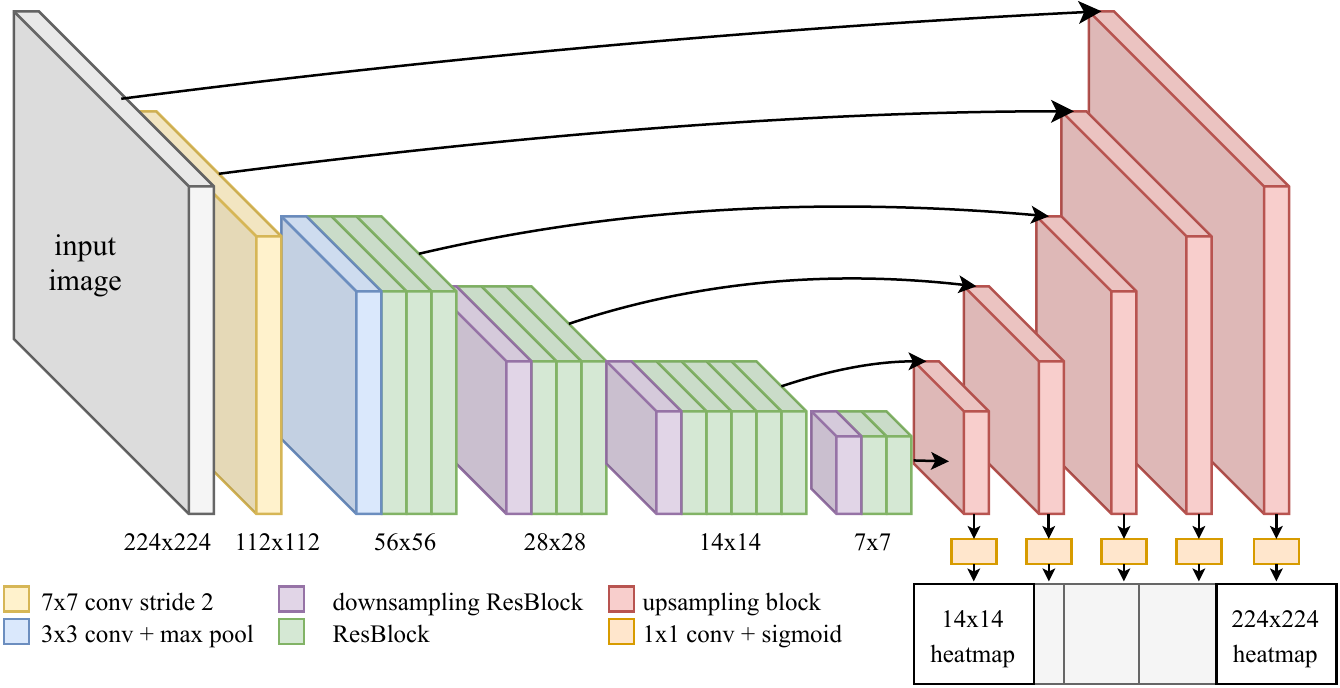}
    \caption{A full overview of the neural network used to generate heatmaps.}
    \label{fig:network_overview}
\end{figure}

A Fully Convolutional Neural Network is trained to generate these heatmaps. To be able to detect features from real images while being trained on synthetic images, a pre-trained feature detector is used. During training the feature detector's layers are frozen, as proposed by \cite{synthetic_hinter}. This is done so that it cannot learn low-level features from the synthetic data as these would not match the features from real images. This has the additional benefit of decreasing the number of required training samples and time. ResNet50~\cite{resnet} is used as it offers good performance due to its relatively small size, while still achieving good accuracy. The network is pretrained on ImageNet~\cite{imagenet}. This is the transfer learning aspect of our approach, as knowledge learned from real images is transferred the synthetic domain.\\
Using consecutive Residual Blocks, ResNet down-samples the $(224, 224, 3)$ input image to a $(7,7, 2048)$ feature map. The resulting feature map is up-sampled five times to generate a feature map of size $(224, 224, f)$ with $f$ the number of filters used. Upsampling happens by using nearest neighbour interpolation to double the width and height of the feature map. Skip connections are used to recover the spatial information that was lost during downsampling to increase the accuracy of the final heatmaps. This is done by concatenating the output of the final activation layer from the corresponding size from the ResNet to the up-sampled feature map. This is followed by two convolutional blocks, consisting of a 2D convolution with kernel size 3, Batch Normalization~\cite{batch_normalization}, a Leaky ReLU and a Dropout~\cite{dropout} layer with a rate of 0.2. An illustration of the upsampling block is shown in Figure~\ref{fig:upsampling}. The lowest level convolutions use 256 filters, this amount is divided by two, each time the feature map is up-sampled. Finally, a convolution with kernel size one, $N$ filters and a sigmoid activation is applied to generate the final heatmaps. $N$ is the desired number of heatmaps.\\
When training neural networks on synthetic data it is desirable that the network uses high level features when generating output as opposed to focusing on lower level features. This is because higher level features, such as shape, match between synthetic and real images; whereas lower level features, such as texture, can differ tremendously. In a classic FCN, output is generated at the final upsampling step. This means that, when using skip connections, output is created from the upsampled feature map combined with the lowest level features from the feature detector network. This could lead to the network lazily focusing on these lower level features, which we are trying to avoid. Intermediate supervision is used to force the network to learn good features that assist in localizing the keypoints, at each level of the upsampling chain. This is implemented by generating a heatmap after each upsampling block and using each heatmap in the loss function. So as opposed to outputting one heatmap, the network generates a set of progressively larger heatmaps, starting at $(14,14)$ up to $(224, 224)$. By applying this type of supervision at each upsampling level, the learning of high level features gets boosted, while lower level features can also still be used to generate the final output. Although, potentially less accurate, each of the intermediate heatmaps can be used to localize keypoints. A full overview of the neural network is given in Figure~\ref{fig:network_overview}.

\subsection{Training}
The loss function is the sum of the Mean Squared Error between the heatmap generated at each level and their corresponding ground truth.
\begin{equation}
    L = \sum_{k} (\frac{1}{n} \sum_{n}(f_k(X_n) - Y_{k,n}))
\end{equation}
Here $f_k$ is the output of the $k$-th upsampling block, $X_n$ the $n$-th image and $Y_{k,n}$ the ground truth heatmaps for the $n$-th image, downscaled to the size of the $k$-th output. Training is done using the ADAM~\cite{adam} optimizer with a learning rate of $1.5e^{-5}$. The learning rate is scaled by $0.1$ if the validation loss has not improved for 10 epochs. A batch size of 16 is used with a maximum of 200 epochs. The network stops training after showing no improvement for 20 epochs. The loss on the validation set is measured at each epoch and the model with the lowest validation error is kept. To improve robustness and generalization towards real life images, a wide range of data augmentation techniques are applied to the images. These include affine transformations, blurring, adding and multiplying, perspective transformations and additive Gaussian noise. When transformations are applied, the heatmaps are adapted correspondingly.

\section{Results}
The proposed method is first validated on a dataset of tool images created specifically for this paper. To show this method generalizes well to other domains and to compare the results of this paper to an established benchmark, the method is also validated on the Pascal3D dataset~\cite{pascal3d}. Our approach has been compared to two other methods for keypoint detection: Stacked Hourglass Networks (SH)~\cite{stacked_hourglass,stacked_github} with two stacks and Simple Baselines (SB)~\cite{simple_baselines}. The former is chosen as it is used often, achieves great accuracy and also employs a form of intermediate supervision. Since this method does not use transfer learning, the latter is also selected for validation, as transfer learning might give an advantage when training on synthetic data. A comparison is also made with a trimmed version of our IHM, that generates heatmaps of size $(56,56)$ ($\textrm{IHM}_{56}$), and with a version of our model that was trained without intermediate supervision~($\textrm{HM}$). \\
The Percentage of Correct Keypoints~(PCK) metric is used to validate performance. A keypoint is considered to be correct when its distance from the ground truth is below a certain fraction of the largest side of the objects bounding box: $ \alpha \cdot max(w, h) $, with $ \alpha $ usually at $0.1$. \\
In the experiments described below, we train a separate neural network for each object, as opposed to one network that can predict the keypoints of each object. The benefit of this approach is that a smaller network is required for each tool, leading to better execution times. For this it is assumed that the system knows which tools keypoints need to be found, as executing the network for each tool would have the opposite effect on the total execution time. Another benefit of this approach is that new objects can easily be introduced into the system by simply training a new network for only that tool. Although training one large network for all objects could also have benefits. Potentially, less training images per object could be needed, as feature learning would be shared, and features could be learned from each object. Additionally this approach could decrease sensitivity to false positives, as the network sees different types of tools during training.

\subsection{Tool Dataset} \label{sec:results_tool}
\begin{figure}
    \centering
    \includegraphics[width=\linewidth]{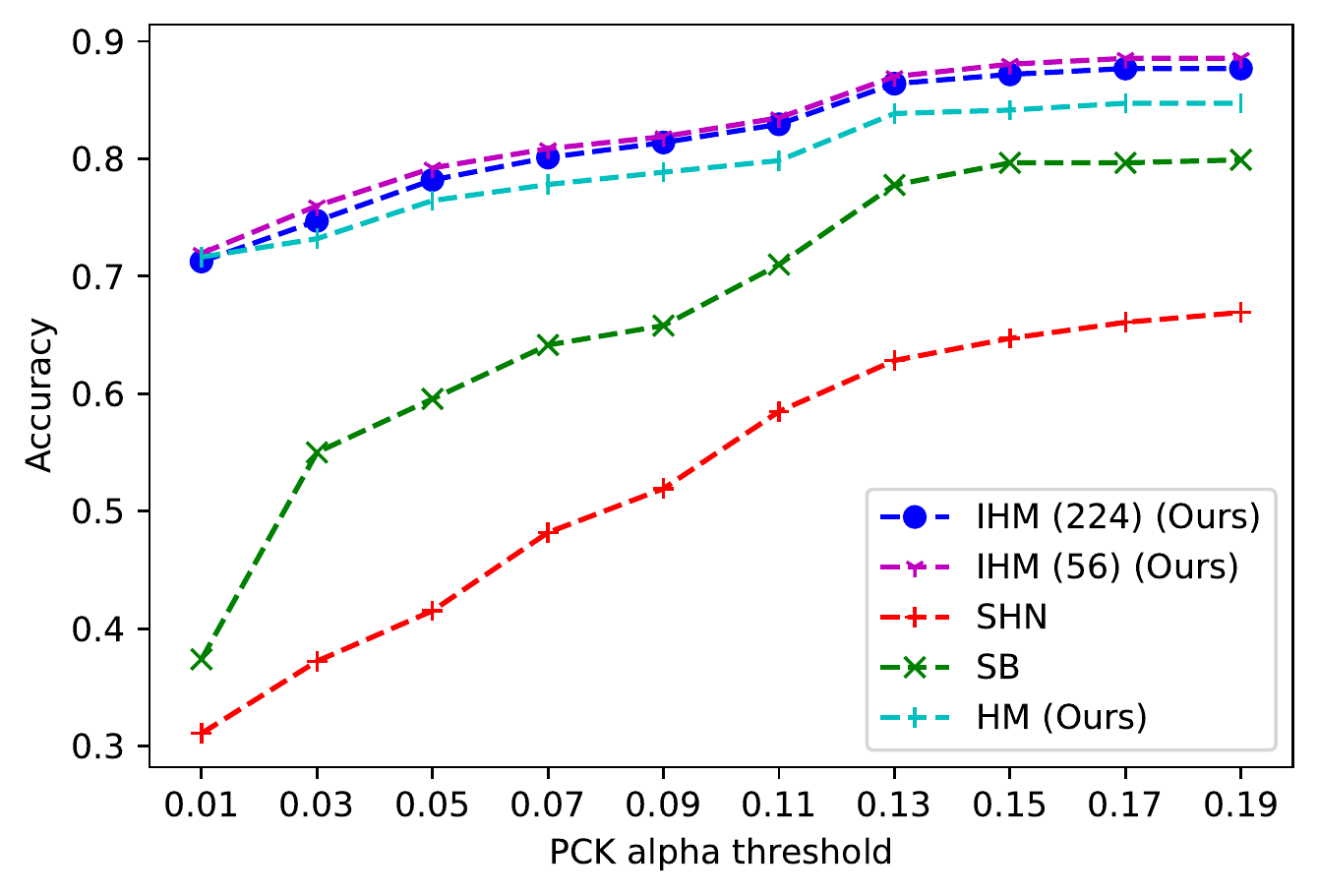}
    \caption{PCK values for the five examined models for varying thresholds.}
    \label{fig:tools_pck}
\end{figure}

\begin{figure}
    \centering
    \includegraphics[width=\linewidth]{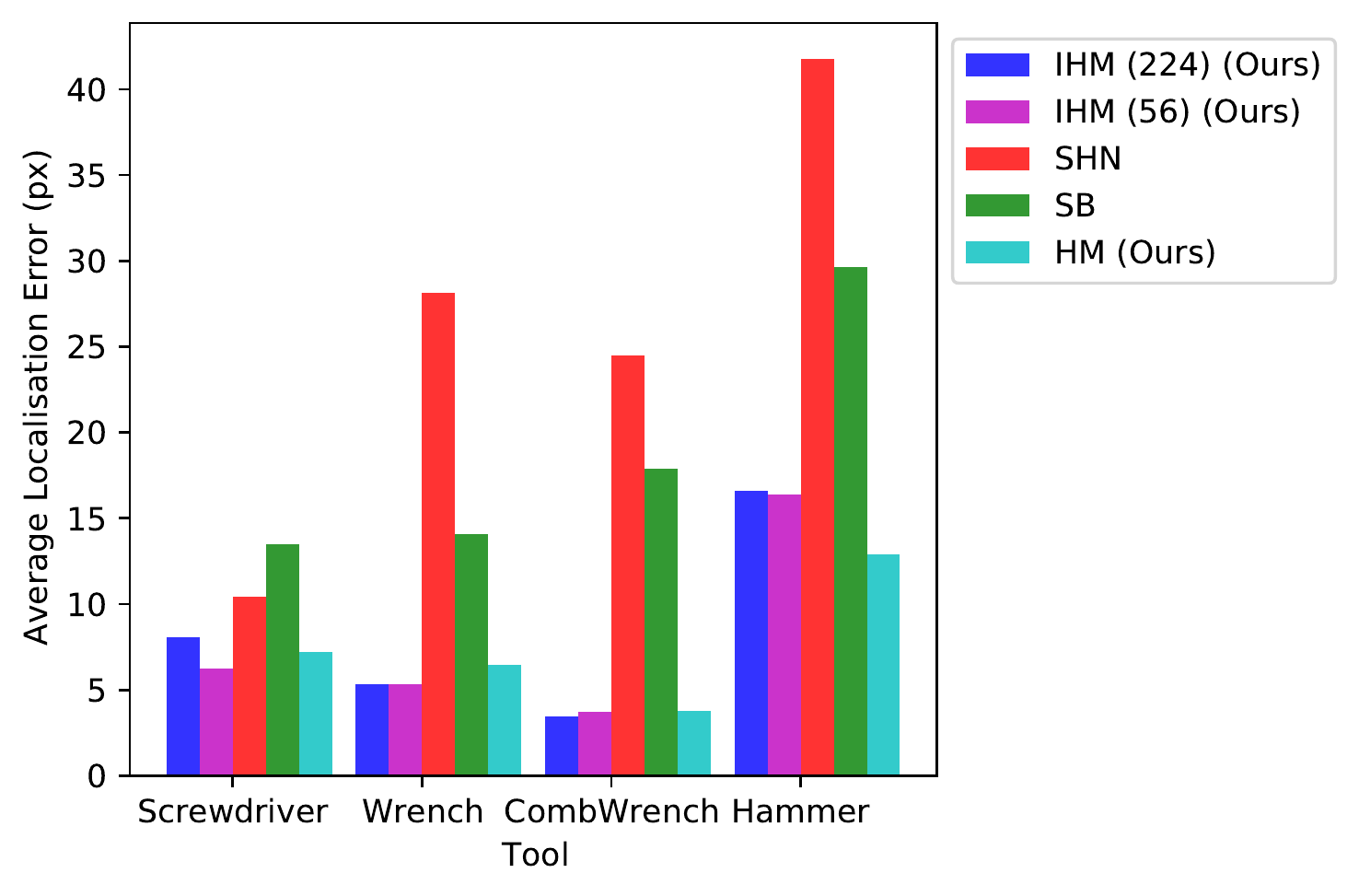}
    \caption{Average localization error for correctly predicted keypoints at $\alpha=0.1$. The images are of size $(224,224)$.}
    \label{fig:error}
\end{figure}

\begin{figure}
\centering
\includegraphics[width = 0.3\linewidth]{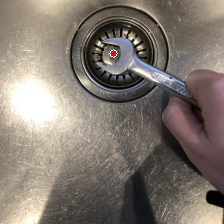}
\includegraphics[width = 0.3\linewidth]{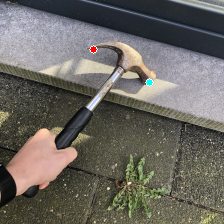}
\includegraphics[width = 0.3\linewidth]{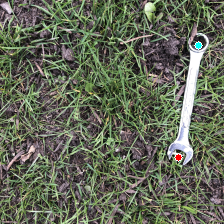}
\includegraphics[width = 0.3\linewidth]{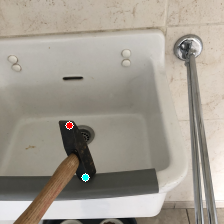}
\includegraphics[width = 0.3\linewidth]{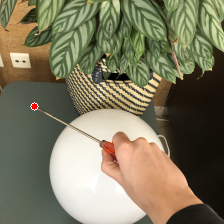}
\includegraphics[width = 0.3\linewidth]{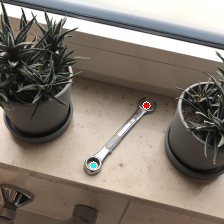}
\caption{Six examples from the tool dataset. Predictions made by the $\textrm{IHM}_{56}$ model are indicated by the dots. Correctly predicted images were chosen.}
\label{fig:tool_dataset}
\end{figure}

To validate our work, a dataset of images has been created for four different tools. Of each tool 50 photographs are created with a mobile phone camera. Photos are taken with a wide range of backgrounds, lighting conditions, camera rotations and positions. The tools are often truncated or partially occluded by hands or other objects. The images are manually annotated with keypoint locations. For each tool six to ten different 3D models with textures were collected from the internet. These models were used to generate a dataset of $10.000$ training images as described in Section~\ref{sec:training_data}. For each tool, the five different models were trained on that dataset.\\
For each model the PCK value was computed on the validation dataset of real images, the results are displayed in Table~\ref{tab:tool_results}. For each of the four tools, the Intermediate Heatmap Model outperforms the Simple Baseline and Stacked Hourglass models. The latter is outperformed by a large margin, highlighting the importance of transfer learning when using synthetic training data. Also the model without intermediate supervision is outperformed for each tool, showing the importance of this learning method. To further examine the accuracy, the performance of the models is measured for differing $\alpha$ values for the PCK. The results, shown in Figure~\ref{fig:tools_pck}, indicate that IHM has a way better localization error, as it manages to still achieve a good PCK score at a very low $\alpha$ value compared to the other models.
\begin{table}[t]
  \caption{Comparison of the IHM, HM, SH and SB methods on the Tool Dataset ($PCK_{\alpha = 0.1}$).} \label{tab:tool_results}
  \begin{center}
    \begin{tabular}{c|p{0.8cm}p{0.8cm}p{0.5cm}p{0.5cm}p{0.5cm}}
      \hline
      \hline
      \textbf{Model} & $\textrm{IHM}_{224}$ & $\textrm{IHM}_{56}$ & HM & SHN & SB\\
       & (Ours) & (Ours) & (Ours) & & \\
      \hline
      Screwdriver   & \textbf{82.0} & \textbf{82.0} & 78.0 & 46.0 & 74.0 \\
      Wrench        & \textbf{88.9} & \textbf{88.9} & 86.1 & 66.7 & 67.6 \\
      C. Wrench     & \textbf{84.9} & \textbf{84.9} & 82.6 & 55.8 & 68.6 \\
      Hammer        & 72.9 & \textbf{74.0} & 68.8 & 45.8 & 63.5 \\
      \hline
      \hline
    \end{tabular}
  \end{center}
\end{table}
Remarkably, the $\textrm{IHM}_{56}$ model with two less upsampling layers performs as well as, or better than the full IHM model. To further examine this, the distance between the keypoints that were correctly predicted with $\alpha = 0.1$ and the actual ground truth is computed. These results per class are shown in Figure~\ref{fig:error}. The $\textrm{IHM}_{224}$ model has an average localization error of 8.4px for correctly predicted keypoints, whereas the $\textrm{IHM}_{56}$ model has an error of 7.9px. The SHN and SB methods have an error of 25.2px and 18.8px respectively. This shows that the final upsampling layers, that utilize the low-level features from the feature detector, offer no benefit to the final accuracy. This indicates the importance of high level features when training on synthetic data. Some correct predictions made by the best performing model $\textrm{IHM}_{56}$ are shown in Figure~\ref{fig:tool_dataset}.
\subsubsection{Performance}
The time needed to process an image is 0.04 seconds on average for all our models, this equates to 25 frames per second. This time includes all post-processing done to find the keypoint locations from the generated heatmaps. This time was measured on a machine with an NVidia GeFore RTX 2080 super. The other models performed within comparable times.

\subsubsection{Cut and Paste vs. 3D generated training data}
\begin{figure}
    \centering
    \includegraphics[width=\linewidth]{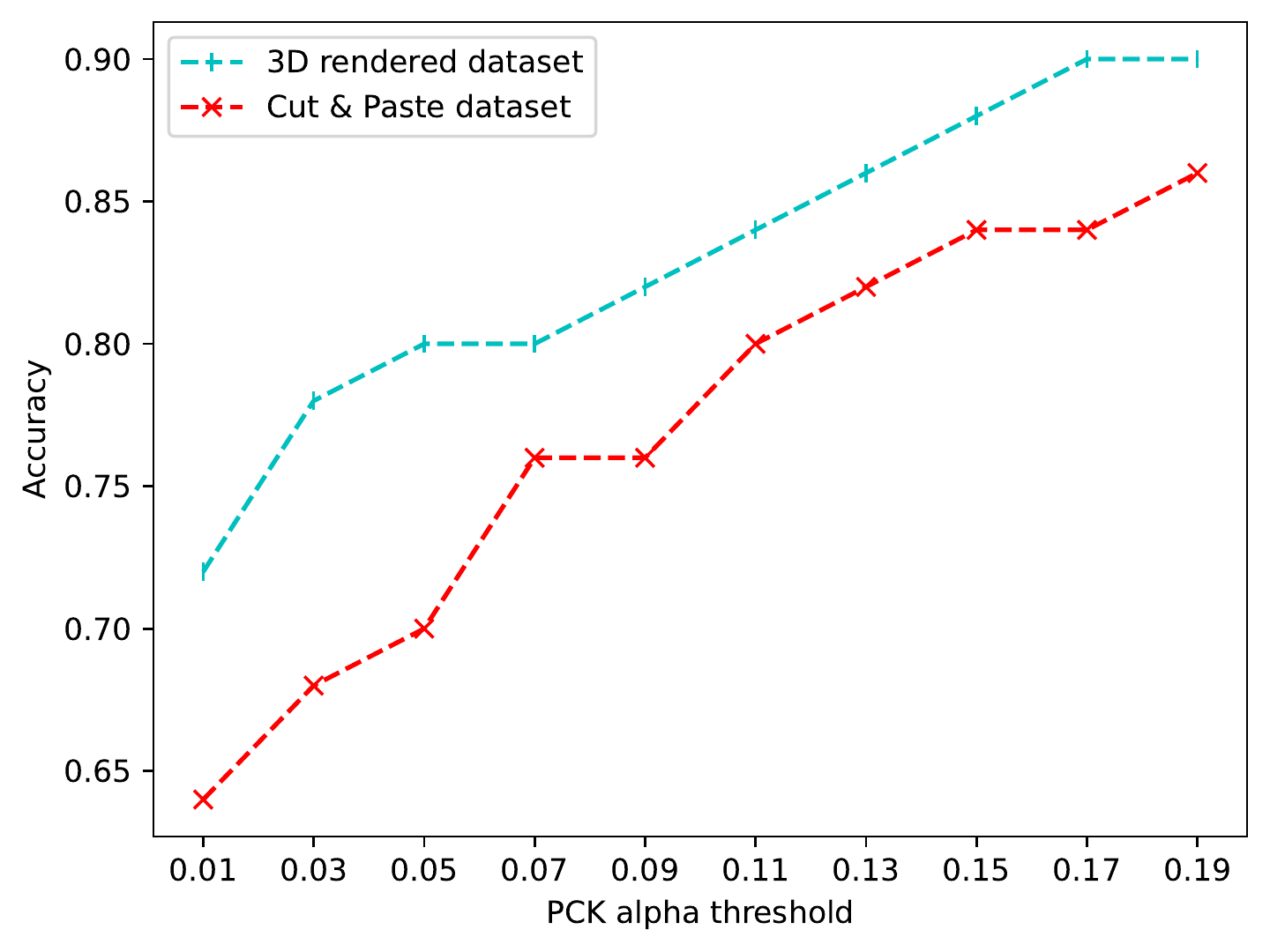}
    \caption{PCK values for varying $\alpha$ thresholds of the $\textrm{IHM}_{56}$ model trained on the Cut and Paste dataset compared to our 3D dataset.}
    \label{fig:3d_vs_2d}
\end{figure}

To show the importance of using 3D generated training images for keypoint localization, as opposed to cut and paste 2D images, an extra experiment is conducted, comparing the two methods. For this experiment the focus is on the screwdriver class. An extra set of images with ground truth annotations is created according to the method of \cite{cut_and_paste}. Images of ten different screwdrivers were collected from the internet and their backgrounds were removed. They were added to random background images with a random rotation, translation and scale using a random blending mode, as described in Section~\ref{sec:training_data} and \cite{cut_and_paste}. This process was repeated to generate 10.000 images. This dataset was then used to train a version of the $\textrm{IHM}$ model. This model is compared to the model trained on the 3D generated data. This comparison for a varying range of $\alpha$ values can be seen in Figure~\ref{fig:3d_vs_2d}. The model trained on 3D data clearly outperforms the model trained on the Cut and Paste generated images. This shows that the extra variation that 3D generated data brings has a clear benefit for the task of keypoint localization.

\subsection{Pascal3D Dataset}
As no established dataset or benchmark exists for this specific problem, the proposed method is also validated on the Pascal3D dataset. Since the results from the previous section already show that our method generalizes to real images when trained on synthetic ones, the focus is now on purely validating the task of keypoint detection. Hence, for the Pascal3D dataset, we train on the real images provided by the dataset and not synthetic ones generated by our method. For each of the classes in the dataset a separate model is trained to detect the keypoints specified in the dataset. A separate validation set is withheld from training and used to compute the PCK values.\\
The results of our models are compared to those of a number of existing methods. This is shown in Table~\ref{tab:pascal_results}. The PCK values are taken from the methods respective papers. For the \cite{semantic_keypoints} method, the PCK value was computed by \cite{starmap}. Our $\textrm{IHM}_{56}$ model achieves an accuracy of 74.2 whereas our HM model achieves 76.7 PCK. Interestingly the model with intermediate supervision performs worse than the model without. This indicates that intermediate heatmap generation improves accuracy when training on synthetic data, but is not of useful when training is done using real data. Our method performs well for most classes, achieving +90\% accuracy for the bus and car classes, but the score is lowered significantly by bad performance on the dining table and train classes. These results show that, although not achieving state-of-the-art, the proposed method still achieves decent performance when applied to a different domain. The focus of this paper is to detect tool keypoints in real-time when trained on synthetic data, so scoring only $5.8$ points below SoA on a different task is reasonable.
\begin{table}[t]
  \caption{Comparison of selected methods on the PASCAL3D Dataset.} \label{tab:pascal_results}
  \begin{center}
    \begin{tabular}{c | c}
      \hline
      \hline
      \textbf{Method} & $\textrm{PCK}_{\alpha=0.1}$\\
      \hline
        $\textrm{IHM}_{56}$ (Ours)     & 74.2 \\
        HM (Ours)                      & 76.7 \\
        Pavlakos et al.~\cite{semantic_keypoints} & \textbf{82.5} \\
        Zhou et al.~\cite{starmap}             & 78.6 \\
        Tulsiani et al.~\cite{viewpoints_and_keypoints}         & 68.8 \\
        Long et al.~\cite{conv_correspondance}             & 48.5 \\
        Simple Baseline~\cite{simple_baselines}         & 42.7 \\
        
      \hline
      \hline
    \end{tabular}
  \end{center}
\end{table}
\section{Conclusion}
In this paper a method for real-time tool keypoint localization was proposed, along with a method for synthetic data generation. Models were successfully trained to detect the landmarks of four different tools in real images, proving the usability of the proposed data generator tool. It was shown that the method outperforms existing methods at the task by predicting the keypoints with a higher accuracy. Additionally the importance of intermediate supervision when using synthetic data was shown, as a copy of our model that does not use this scheme performs worse on synthetic data. Furthermore, we have shown that an exact 3D model of an object is not necessary when generating synthetic training images, as long as a set of varying 3D models are used. Validation on the Pascal3D dataset has shown that this method also generalizes to domains other than tools.

\vfill
\section*{\uppercase{Acknowledgements}}
This study was supported by the Special Research Fund (BOF) of Hasselt University. The mandate ID is BOF20OWB24. Research was done in alignment with Flanders Make's PILS and FAMAR projects.

\bibliographystyle{apalike}
{\small
\bibliography{bibliography}}

\end{document}